\documentclass[10pt,twocolumn,letterpaper]{article}

\usepackage{cvpr}
\usepackage{times}
\usepackage{epsfig}
\usepackage{graphicx}
\usepackage{amsmath}
\usepackage{amssymb}

\usepackage[breaklinks=true,bookmarks=false]{hyperref}

\cvprfinalcopy % *** Uncomment this line for the final submission

\def\cvprPaperID{****} % *** Enter the CVPR Paper ID here

\begin{document}

%%%%%%%%% TITLE
\title{SCR-Graph: Spatial-Causal Relationships based Graph Reasoning Network for Human Action Prediction}

\author{Bo Chen\\
Chinese Academy of Sciences\\
Shenyang, Liaoning\\
{\tt\small edentime@163.com}
\and
Decai Li\\
Chinese Academy of Sciences\\
Shenyang, Liaoning\\
{\tt\small lidecai@sia.cn}
\and
Yuqing He\\
Chinese Academy of Sciences\\
Shenyang, Liaoning\\
{\tt\small heyuqing@sia.cn}
\and
Chunsheng Hua\\
Liaoning University\\
Shenyang, Liaoning\\
{\tt\small huachunsheng@lnu.edu.cn}
}

\maketitle
\thispagestyle{empty}

%%%%%%%%% ABSTRACT
\begin{abstract}

Technologies to predict human actions are extremely important for applications such as human robot cooperation and autonomous driving. However, a majority of the existing algorithms focus on exploiting visual features of the videos and do not consider the mining of relationships, which include spatial relationships between human and scene elements as well as causal relationships in temporal action sequences. In fact, human beings are good at using spatial and causal relational reasoning mechanism to predict the actions of others. Inspired by this idea, we proposed a Spatial and Causal Relationship based Graph Reasoning Network (SCR-Graph), which can be used to predict human actions by modeling the action-scene relationship, and causal relationship between actions, in spatial and temporal dimensions respectively. Here, in spatial dimension, a hierarchical graph attention module is designed by iteratively aggregating the features of different kinds of scene elements in different level. In temporal dimension, we designed a knowledge graph based causal reasoning module and map the past actions to temporal causal features through Diffusion RNN. Finally, we integrated the causality features into the heterogeneous graph in the form of “shadow node”, and introduced a self-attention module to determine the time when the knowledge graph information should be activated. Extensive experimental results on the VIRAT datasets demonstrate the favorable performance of the proposed framework.

\end{abstract}

%%%%%%%%% BODY TEXT
\section{Introduction}

In recent years, the interest in human action prediction is increasing owing to its broad and important applications such as autonomous driving \cite{bhattacharyya2018long}, human-robot cooperation \cite{mainprice2016goal},and security monitoring \cite{liang2019peeking}.

\begin{figure}[t]
\begin{center}
\includegraphics[width=1.00\linewidth]{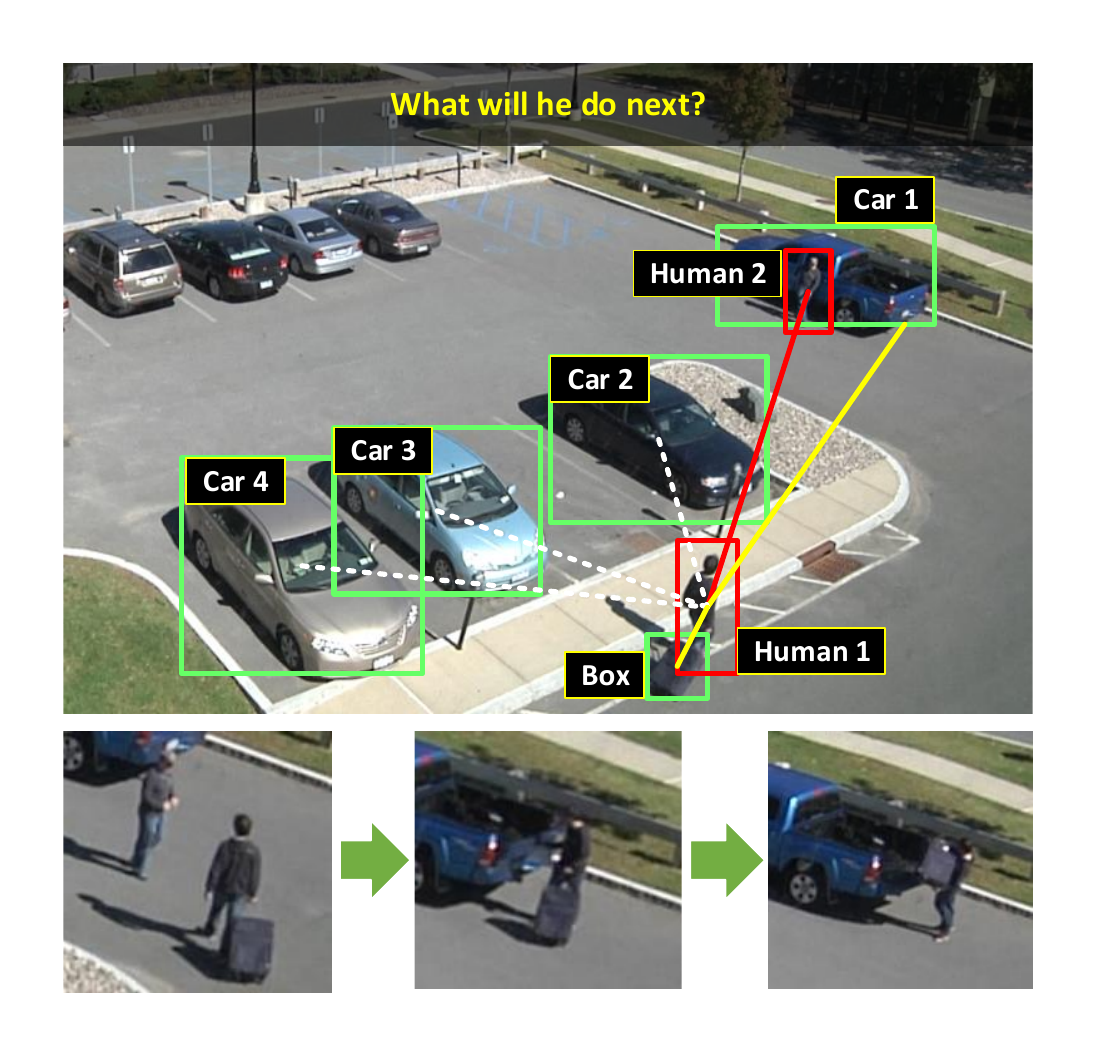}
\end{center}
   \caption{What determines a human's future actions?}
\label{img1}
\end{figure}

However, human action prediction is a particularly challenging problem owing to three main reasons (Figure.\ref{img1}): (1) Future actions of humans are internally driven with a certain purpose; however, because this purpose is not visible, we can only make inferences and judgments based on available external information \cite{zlotowski2015anthropomorphism}. (2) The change of surrounding environment or the intervention of other people affects the next step in human action, by affecting the decision-making \cite{alahi2016social}. (3) Even if human current action is the same in similar environment, different past actions can cause different action choices to be made in the future. Below, we will briefly review the relative works about human action prediction and introduce SRC-Graph method that proposed in this paper.

In recent years, various attempts have been made in human action prediction. The works in \cite{ryoo2011human} \cite{li2014prediction}\cite{hoai2014max} identified activity prediction with early detection of short-duration single action using hand-crafted features; however, the limited expression ability of these features limits the effect of action prediction. With the rapid development of deep learning, researchers began to use deep neural network (two stream CNNs \cite{feichtenhofer2016convolutional}, RNNs \cite{wu2015modeling}, 3D-CNNs \cite{tran2015learning}\cite{qiu2017learning}, etc.) to automatically extract video features, and predict actions based on these features. \cite{kong2018adversarial} proposed a temporal deep model to better learn activity progression for performing activity detection and early detection tasks. But in this case, it is easy to be disturbed by sample noise due to the model’s inability to extract the key information of video accurately. \cite{kong2018action} designed a mem-LSTM model using CNN and LSTM to model the spatial and time dimensions. However, because of taking the raw video images as the input  and ignoring mining the relationships behind, the prediction performances of these methods are limited in both accuracy and time horizon.

Recently, \cite{bhattacharyya2018long} proposed a model that pays particular attention to the subjects in the videos with human detection and tracking over large time horizons. Good predictions have been made in predicting the person's behavior for the next second, so it cannot predict the human actions after a longer time ( 5 to 10 seconds). Li Fei-Fei and her team proposed an end-to-end, multi-task learning system utilizing rich visual features \cite{liang2019peeking}. They encoded a person by using rich semantic features about visual appearance, body movement and the person’s interaction with the surroundings. Their study was motivated by the fact that humans derive such predictions by relying on similar visual cues, and it achieved remarkable better results and predict the future 4.8 seconds (12 frames) of person trajectory,

However, the methods above mainly focus on adding together, the visual features about the human behavioral information, and the interaction of humans with their surroundings directly. The influence of the relationship between different objects and pedestrians on the future behavior of humans in the scene was not deeply explored. In addition, they did not consider the causality in action sequences. For example, as shown in Figure.\ref{img1}, if a person tend to a car while pulling a box , in the next moment, he will have exhibit a high probability of opening the trunk to place the box and then opening the door to get into the car.

In consideration of the above factors, in order to accurately predict the actions in a long period of time, we need to not only model the relationship between human behaviors and objects in the scene, but also to conduct the causal reasoning of transformation of actions according to the time series of the actions. 

We proposed a spatial and causal reasoning graph network (SCR-Graph) to model the surrounding relationships in spatial dimension and the causal relationships in time dimension. This is structurally composed of two parts: The first part in spatial dimension is designed to capture the features of spatial topological relation between the subject and the surrounding environment elements. This is achieved by constructing a hierarchical heterogeneous graph neural network, so as to obtain the relation influence of the surrounding objects and others on the subject. The second part in time dimension is proposed to create a causal reasoning module, whose function is to constrain the action prediction results by using the action causal relationship reasoning, to make it fit the logical relationship between the action sequences. At the same time, considering that the switching time of actions is affected by multiple factors, we integrated the causality features (which were acquired by the Gate-Graph reasoning on knowledge graph in the time dimension) into the heterogeneous graph in the form of “shadow node”, and introduced a self-attention module to judge when the knowledge graph information should be activated during processing.

\textbf{The main contributions of this study are summarized as follows:}

(1) Inspired by the principles of humans’ powerful action prediction ability, we proposed a two-stream graph framework called SCR-Graph. Through construction of the surrounding relationships modeling module in spatial dimension and the causal reasoning module in time dimension, this model acquired the ability of logical reasoning in two dimensions, thus imitating humans.

(2) To improve the model’s performance and generalization ability, we construct Diffusion-RNN as a guidance to automatically determine the action prediction results. We further integrated the causality features into the heterogeneous graph in the form of “shadow node”.

(3)	We conducted extensive experiments on action prediction datasets and demonstrated the effectiveness of the proposed framework.

%-------------------------------------------------------------------------
\section{Related work}

The introduction of unstructured data modeling and processing methods has greatly improved our algorithm. In this section, we briefly review the existing methods on two problems related to our work: graph neural networks and knowledge graphs.

\begin{figure*}
\begin{center}
\includegraphics[width=17.5cm]{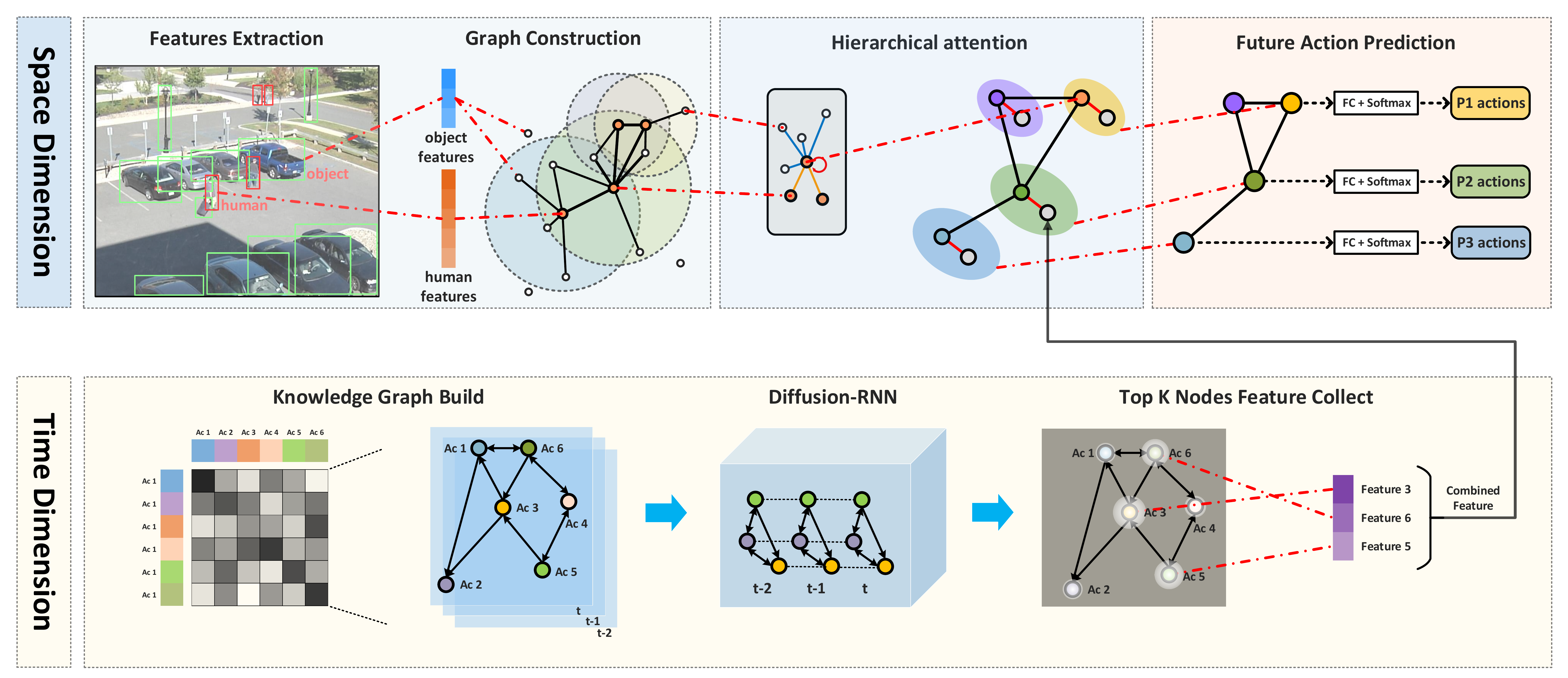}
\end{center}
   \caption{ The overall framework of SCR-Graph.}
\label{img2}
\end{figure*}

\subsection{Graph Neural Networks.}

Graph neural networks (GNNs) are connectionist models that capture the dependence of graphs via message passing between the nodes of graphs \cite{zhou2018graph}. These can be divided into two types: graph convolution neural networks (GCN) and graph gate neural networks (Gate-GNN).

For graph convolution neural network, advances in this direction are often categorized as spectral approaches and non-spectral approaches. \cite{bruna2013spectral} proposed the spectral network, whose convolution operation is defined in the Fourier domain by computing the eigen decomposition of the graph Laplacian. But a model trained on a specific structure could not be directly applied to a graph with a different structure, which limited their application scenarios. Non-spectral approaches define convolutions directly on the graph operating on spatially close neighbors \cite{duvenaud2015convolutional}. \cite{hamilton2017inductive} proposed the GraphSAGE, a general inductive framework, which can be applied to different graph structures as it generates embeddings by sampling and aggregating features from a node’s local neighborhood. \cite{velivckovic2017graph} proposed a graph attention network (GAT) which incorporates the attention mechanism into the propagation step. It computes the hidden states of each node by attending over its neighbors, following a self-attention strategy. And \cite{wang2019heterogeneous} extends attention mechanism to heterogeneous graphs to deal with those which contain different types of nodes and links.

Gate based graph neural network attempts to use the gate mechanism, similar to GRU \cite{cho2014learning} or LSTM \cite{hochreiter1997long}, in the propagation step to diminish the restrictions in the former GNN models and improve the long-term propagation of information across the graph structure \cite{zhou2018graph}.\cite{seo2018structured} combined Graph Convolutional Networks (GCNs) and RNN to model spatial structures and dynamic patterns. Although this achieves the reasoning function of graph structure information in time dimension, but it cannot be used in directed graphs. To model the traffic flow as a diffusion process on a directed graph, \cite{li2017diffusion} proposed Diffusion Convolutional Recurrent Neural Network (DCRNN), which has the ability to capture the spatiotemporal dependencies on directed graphs.

\subsection{Knowledge Graphs}

Knowledge graph is a kind of structured knowledge base, which describes the concepts, entities and their relationships in the objective world in a structured way \cite{kejriwal2019domain}. It can be applied to various kind of tasks, such as situation recognition \cite{li2017situation}, object detection \cite{fang2017object} and visual relationship extraction \cite{lu2016visual}. 

A knowledge based framework which can generalize to other tasks was proposed in \cite{zhu2015building}. \cite{marino2016more} introduced a graph search neural network (GSNN), which can exploit large knowledge graphs into an end-to-end framework for image classification \cite{gao2018watch}, this was different from the methods which treat knowledge graph as a separate component in their frameworks. In addition, \cite{gao2018watch} applies knowledge graphs for video classification, and propose a novel knowledge-based attention model. However, using knowledge graph to solve the problem of human action prediction has not been studied in depth yet.

\section{Our Approach}

\textbf{Problem Definition: }We assumed that we already know the tracking bounding box of all people and objects in the scene, based on which we can obtain the features of people and objects easily. Given continuous video frames with pedestrian tracking bounding boxes and object detection results, our model aims to predict each person's actions in the frame in the future.

\textbf{Overall Framework: }Fig.\ref{img2} shows the overall network architecture of our SC-Graph model. 

\textbf{(a)} In spatial dimension: We constructed a hierarchical heterogeneous graph attention network (H-GAT) to get the interactive intention features of human-human and human-scene interactions through the topological relationship between humans and objects in the scene. In addition, to make the network capable of introducing causal reasoning features when necessary, and for better integration of time dimension features, we designed a shadow node module with fusion weight self-adjusting function.

\textbf{(b)} In time dimension: Based on the principle of statistics, we construct Diffusion-RNN that have the ability of graph causal reasoning to get the features and scores of different action nodes in the next moment. Then, we fuse the features of the top K nodes and send them to the shadow nodes to guide the results of action prediction more logically.

\subsection{Graphs Construction methods}

The construction of graph structure is an important foundation of graph neural network reasoning. The quality of graph structure construction determines the subsequent reasoning effect.

\begin{figure}[h]
\begin{center}
\includegraphics[width=0.95\linewidth]{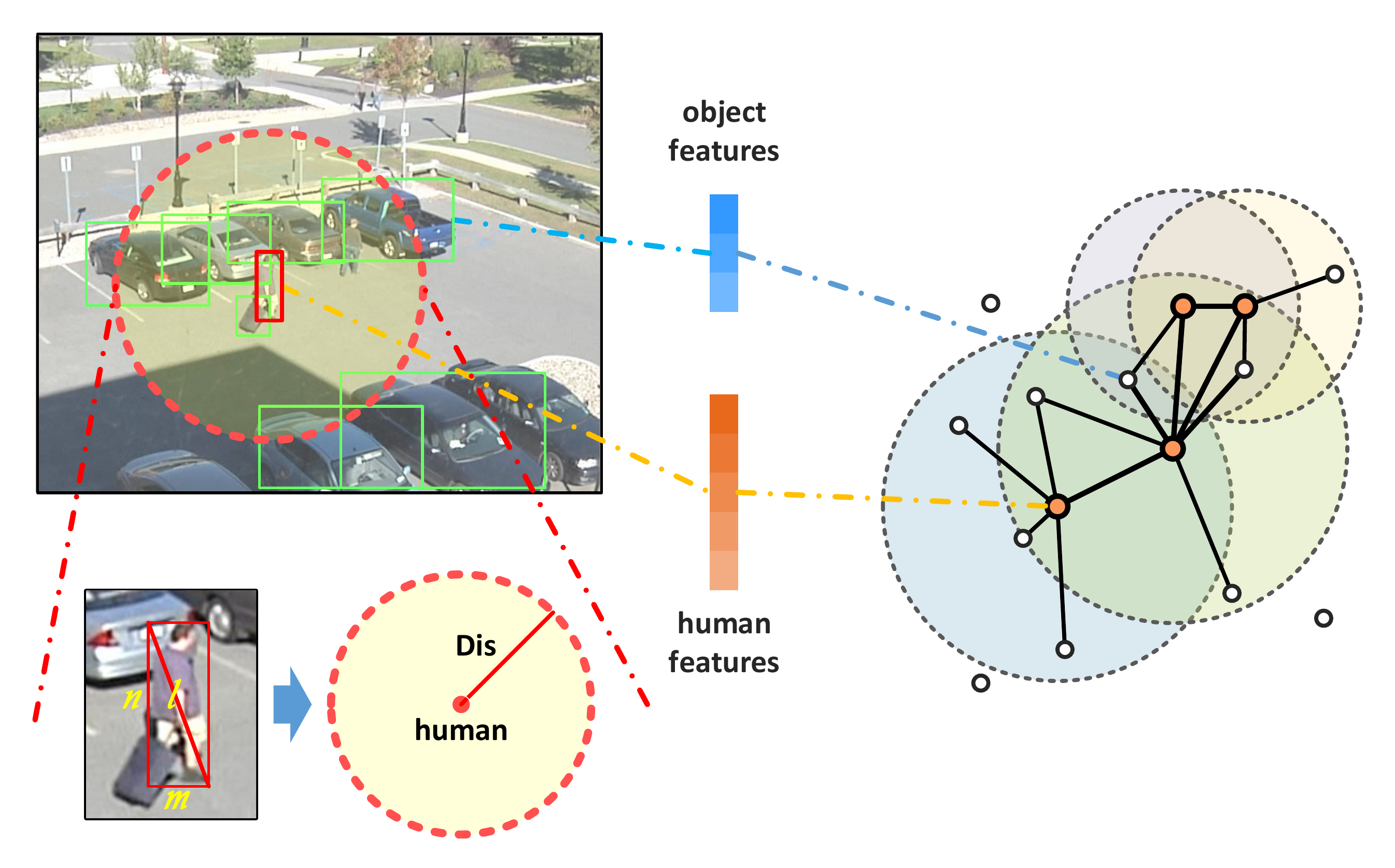}
\end{center}
   \caption{Spatial relationship graph construction.}
\label{img3}
\end{figure}

As shown in Figure.\ref{img3}, the targets that may attract human interaction intention include people and objects in the scene. Therefore, building the relationship between humans and objects is a critical issue. What comes first is to reduce the influence of irrelevant people and objects on a human's action prediction in the complex environment. Here, we set up a human's behavior perception range circle according to each person's size in the picture. The radius $Dis_{(id)}$ of this circle is determined using Formula \ref{eq1} and \ref{eq2}:

\begin{equation}
\label{eq1}
l_{(id)}=\sqrt{{m_{(id)}}^2+{n_{(id)}}^2}
\end{equation}

\begin{equation}
\label{eq2}
Dis_{(id)}=\lambda \cdot l_{(id)}
\end{equation}

Where $m_{(id)}$, $n_{(id)}$ and $l_{(id)}$ are the width, height and diagonal length of pedestrian No. $id$ in the picture respectivel, and $\lambda$ is a constant coefficient. 

Then, we connected the objects within the subject's perception scope with the subject. But considering the special influence of other people on the subject's interaction intention in the scene (for example, another person far away from the scene wave his hand to subject, which will also cause this subject's interaction intention), we constructed the graph structure in the form of full connection for humans. Using this rule, we can obtain the spatial dimension graph structure in any scene.

\begin{figure}[h]
\begin{center}
\includegraphics[width=0.95\linewidth]{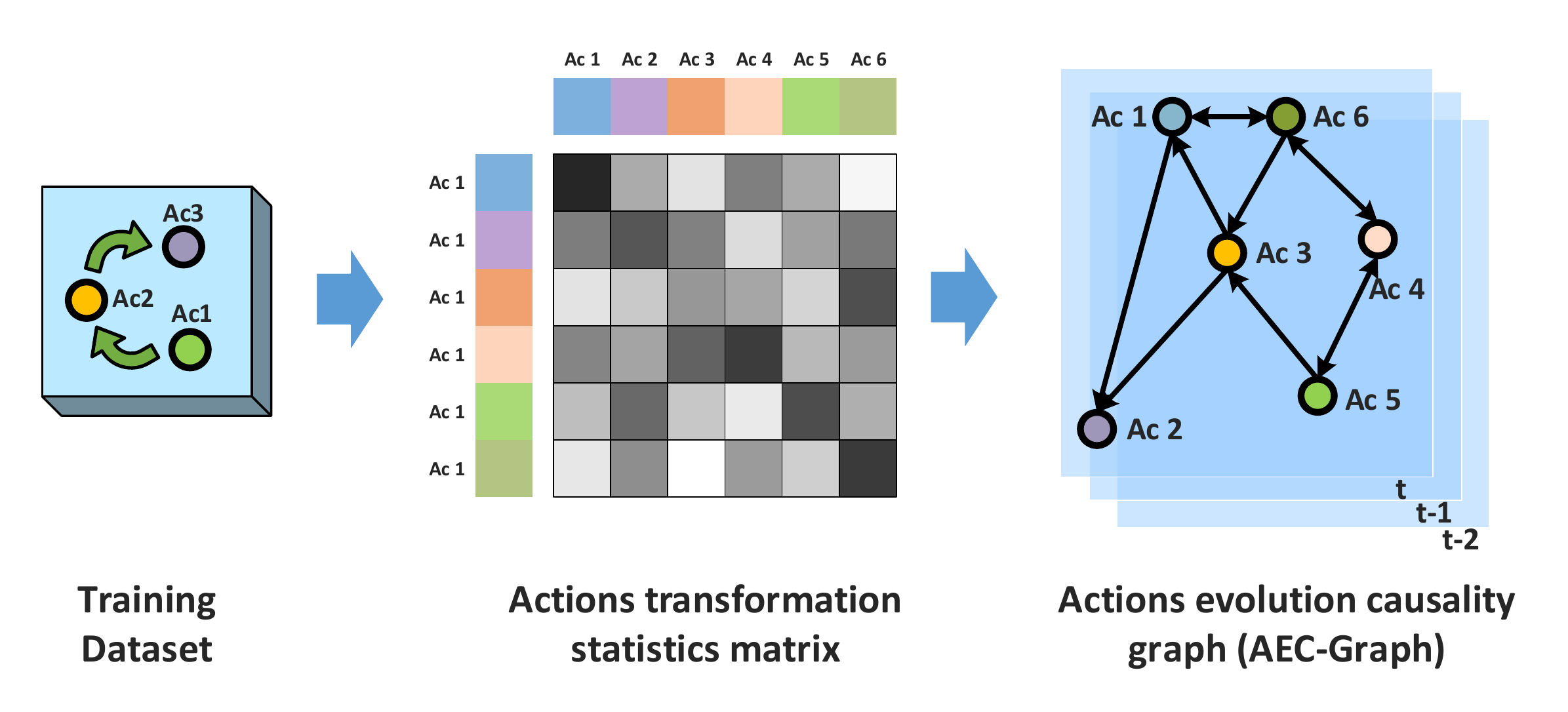}
\end{center}
   \caption{Temporal causal relationship graph construction.}
\label{img4}
\end{figure}

In time dimension, as the causality of actions are directed relation, constructing a directed knowledge graph can obtain better experimental results and make the model more convergent. The knowledge graph is a causality transfer graph between different actions which is obtained by analyzing the actions transfer of training dataset with statistical method. As shown in Figure.\ref{img4}, we calculate the transition probability matrix $\textbf{W}$ of actions first, then build the weighted dirrected graph structure representing the actions switching relationships (more details is explaned in \textbf{Section 4.3}).

\subsection{Action-scene relationship reasoning in spatial dimension}

After got the spatial relation graph structure, the next question is how to define and extract the features of each type of node. For "human" type nodes, the features that may have constraints on future actions include apparent features (human appearance features, semantic features of human-scene, etc.) and motion features (human trajectory features, skeleton point motion features, etc.). Here we use LSTM to encode all the human node related features, which is same to \cite{liang2019peeking}. And for “object” type nodes, the features contain objects’ catageray, size and location informations.

\begin{figure}[h]
\begin{center}
\includegraphics[width=1.00\linewidth]{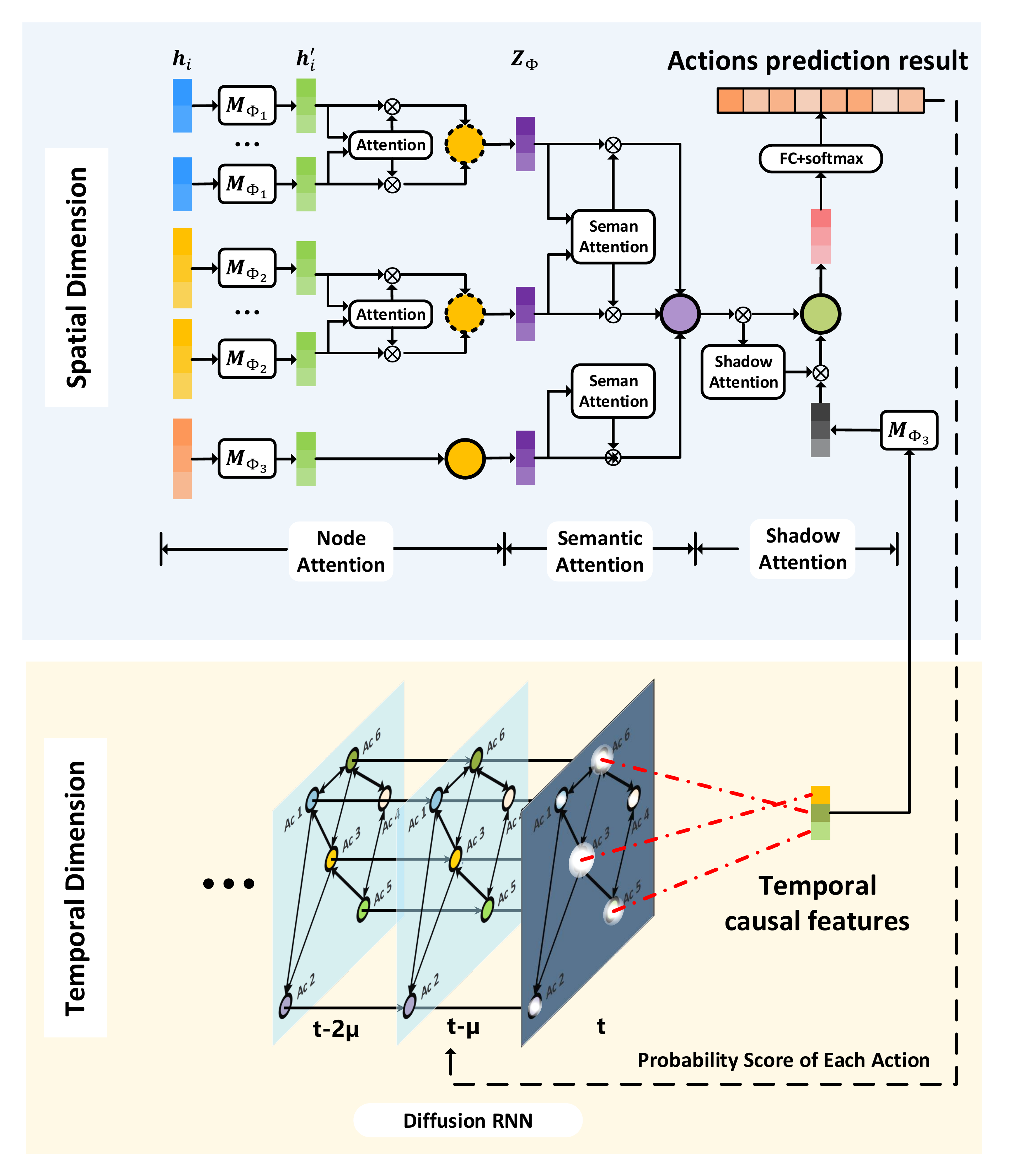}
\end{center}
   \caption{Action-scene hierarchical features aggregation process.}
\label{img5}
\end{figure}

As shown in Figure.\ref{img5}, the hierarchical features aggregation process on the graph can be divided into three steps: node level, type level and shadow level aggregation. First of all, node level aggregation is the process of aggregating node features of the same type. As each node play a different role and show different importance in learning node embedding for the specific task, we introduce node-level attention can learn the importance of same type neighbors for each node and aggregate the representation of these meaningful neighbors to form a node embedding \cite{wang2019heterogeneous}.  

Due to the heterogeneity of nodes, different types of nodes have different feature spaces. Therefore, for each type of nodes, we design the type-specific transformation matrix $\textbf{M}_{\phi_{i}}$ to project the features of different types of nodes into the same feature space (Formula \ref{eq3}).

\begin{equation}
\label{eq3}
\textbf{h}_{i}^{'}=\textbf{M}_{\phi_{i}} \cdot \textbf{h}_i
\end{equation}

Where $\textbf{h}_i$ and $\textbf{h}_{i}^{'}$ are the original and projected feature of node $i$ , respectively. Then we leverage self-attention to learn the weight among various kinds of nodes. The importance of node pair $(i, j)$ in node level can be formulated as Eq. \ref{eq4} and get the weight coefficient $e_{ij}^{\Phi}$ via softmax function as Eq. \ref{eq5}

\begin{equation}
\label{eq4}
e_{ij}^{\Phi}=att_{node}(\textbf{h}_{i}^{'}, \textbf{h}_{j}^{'}; \Phi)
\end{equation}

\begin{equation}
\label{eq5}
\alpha_{ij}^{\Phi}=softmax_{j} (e_{ij}^{\Phi})
\end{equation}

Then, the embedding of node $i$ in each type can be aggregated by the neighbor’s projected features with the corresponding coefficients as follows (Eq. \ref{eq6}):

\begin{equation}
\label{eq6}
z_{i}^{\Phi}=\mathop{\parallel}\limits_{k=1}^{K}\sigma(\sum_{j \in N_{i}^{\Phi}} \alpha_{ij}^{\Phi} \cdot \textbf{h}_{j}^{'})
\end{equation}

Where $z_{i}^{\Phi}$ is the learned embedding of node $i$ for the meta-path $\Phi$. And we extend node-level attention to multihead attention ($K$ is the head number) so that the training process is more stable.

For the type-level attention, the importance of each meta-path, denoted as $w_{\Phi_{i}}$, is shown as follows (Eq. \ref{eq7}):

\begin{equation}
\label{eq7}
w_{\Phi_{i}}=\frac{1}{|V|} \sum\limits_{i \in V}\textbf{q}^T \cdot tanh(\textbf{W} \cdot \textbf{z}_{i}^{\Phi}+\textbf{b})
\end{equation}

Here, $\textbf{W}$ is the weight matrix, $\textbf{b}$ is the bias vector, $\textbf{q}$ is the semantic-level attention vector. The weight of meta-path ${\Phi_{i}}$, denotedas $\beta_{\Phi_{i}}$,can be obtained by normalizing the above importance of all meta-paths using softmax function as Eq. \ref{eq8}:

\begin{equation}
\label{eq8}
\beta_{\Phi_{i}}=\frac{exp{(w_{\Phi_{i}})}}{\sum_{i=1}^{P}exp{(w_{\Phi_{i}})}}
\end{equation}

With the learned weights as coefficients, we can fuse these semantic-specific embeddings to obtain the embedding $\textbf{Z}_c$ as follows (Eq. \ref{eq9}):

\begin{equation}
\label{eq9}
\textbf{Z}_c=\sum_{i=1}^{P} \beta_{\Phi_{i}} \cdot \textbf{Z}_{\Phi_{i}}
\end{equation}

Finally, in order to achieve the function of autonomously fusing the output features of causal inference at the right time, we introduce shadow nodes into the model. The aggregation function in shadow level is formulate by Eq. \ref{eq10}:

\begin{equation}
\label{eq10}
\textbf{Z}_o=\sigma(\textbf{Z}_c+att_{node}(\textbf{Z}_c;\Phi) \cdot \textbf{Z}_s)
\end{equation}

In which $\textbf{Z}_s$ is the shadow node features and $\textbf{Z}_o$ is the final embeded node features for action predicting. After processed by FC and softmax module, the action prediction results can be obtained. As it is a multi-lables classification problem, we choose the BEC loss which is designed as:

\begin{equation}
\label{eq11}
\textbf{L}(x,y)=\sum_{n=0}^{N}{-\omega_{n} [y_n \cdot \log\sigma(x_n)+(1-y_n) \cdot \log (1-\sigma(x_n))]}
\end{equation}

Here, $x$ and  $y$ are the prediction result and ground truth, respectively. $N$ is the number of categories of actions and $\omega_n$ is the weight.

\subsection{Causal relationship reasoning module in temporal dimension}
As there are many causal relationships in human actions sequence (such as upload box then get in the trunk), and people often have multiple actions at the same time (such as talk while walking). How to model the multiple co-occurrence actions and causal reasoning in temporal dimension is the focus of this part.

We can represent the actions switching causality graph as a weighted directed graph $G=(V, E, \textbf{W})$, where $V$ is a set of nodes $|V|=N$, $E$ is a set of edges and $\textbf{W}\in\mathbb{R}^{N \times N}$ is a weighted adjacency matrix representing the nodes proximity \cite{li2017diffusion}.

\textbf{Spatial dependency modeling: }

In order to model the relationship between co-occurrence behaviors, we introduce the diffusion convolution into our method. The diffusion process can be represented as a weighted combination of infinite random walks on the graph \cite{teng2016scalable}.

The resulted diffusion convolution operation over a graph signal $\textbf{X}\in\mathbb{R}^{N \times P}$ and a filter $f_{\theta}$ is defined as:

\begin{equation}
\label{eq12}
\textbf{X}_{:,p}\star{f_\theta}=\sum_{k=1}^{K-1}(\alpha(1-\alpha)^k(\textbf{D}_{O}^{-1}\textbf{W})^k)\textbf{X}_{:,p} \quad for \quad p \in 1,...,P
\end{equation}

Here, $\theta\in\mathbb{R}^K$ are the parameters for the filter and $\textbf{D}_{O}^{-1}\textbf{W}$ represent the transition matrice of the diffusion process. In which, $\textbf{D}_O=diag(\textbf{Wl})$ is the out-degree diagonal matrix and $l$ is the all one vector. $P$ is the dimension of each node's feature \cite{li2017diffusion}.

With the convolution operation defined in Eq.\ref{eq12}, we can build a diffusion convolutional layer that maps P-dimensional features to Q-dimensional outputs. The diffusion convolutional layer is formulated by Eq.\ref{eq13}:

\begin{equation}
\label{eq13}
\textbf{H}_{:,q}=\textbf{a}(\sum_{p=1}^P\textbf{X}_{:,p}\star{f\Theta_{q,p,:,:}}) \quad for \quad \in 1,...,Q
\end{equation}

where $\textbf{H}_{:,q}$ is the output,$\{{f\Theta_{q,p,:,:}}\}$ are the filters and $\textbf{a}$ is the activation function. 

\textbf{Temporal dynamics modeling: }

We leverage the Gated Recurrent Units (GRU) \cite{chung2014empirical} to model the temporal dependency. The Diffusion Convolutional Gated Recurrent Unit (DCGRU) is formulated by Eq.\ref{eq14}-\ref{eq17}.

\begin{equation}
\label{eq14}
\textbf{r}^{(t)}=\sigma(\Theta_r \star G[\textbf{X}^{(t)}, \textbf{H}^{(t-1)}]+\textbf{b}_r)
\end{equation}

\begin{equation}
\label{eq15}
\textbf{u}^{(t)}=\sigma(\Theta_u\star G[\textbf{X}^{(t)}, \textbf{H}^{(t-1)}]+\textbf{b}_u)
\end{equation}

\begin{equation}
\label{eq16}
\textbf{C}^{(t)}=tanh(\Theta_C \star G[\textbf{X}^{(t)}, (\textbf{r}^{(t)}\odot \textbf{H}^{(t-1)})]+\textbf{b}_c)
\end{equation}

\begin{equation}
\label{eq17}
\textbf{H}^{(t)}=\textbf{u}^{(t)} \odot \textbf{H}^{(t-1)}+(1-\textbf{u}^{(t)}) \odot \textbf{C}^{(t)}
\end{equation}

where $\textbf{X}^{(t)}$, $\textbf{H}^{(t)}$denote the input and output of at time $t$ , r$\textbf{r}^{(t)}$, $\textbf{u}^{(t)}$ are reset gate and update gate at time $t$, respectively. $\Theta_r$, $\Theta_u$, $\Theta_C$ are parameters for the corresponding filters.

\section{Experiments}

In this section, we evaluate the performance of the proposed method on the ActEV/VIRAT  \cite{awad2018trecvid}\cite{oh2011large} dataset for action prediction. The extensive results thus obtained demonstrate the effectiveness of our method for the prediction of human actions. Finally, we conducted a component analysis of our framework according to several examples.

\subsection{Implementation Details}

\textbf{Dataset: }ActEV/VIRAT \cite{awad2018trecvid} is a public dataset released by NIST in 2018 for activity detection research in streaming video. This dataset is an improved version of VIRAT \cite{oh2011large} , with more videos and annotations. It includes 455 videos at 30 fps from 12 scenes, more than 12 hours of recordings. Most of the videos have a high resolution of 1920x1080.

\textbf{Features: }All the pre-extracted features used in this model, including: human keypoints motion feature \cite{he2017mask}, human appearance feature \cite{lin2017feature}, trajectory feature, human-object spatial feature and human-scene feature, are extracted through pre-trained model and uniformly encoded through LSTM \cite{liang2019peeking}. Each feature is encoded as a vector of length 256.

\subsection{Results and Analysis}

\noindent\textbf{(1) Knowledge graph construction: }

We tested the construction effect of knowledge graph first, since whose quality has a great influence on the subsequent graph reasoning effect. 

We use statistical method to count the actions switching relationship and calculate the switching probability. In details, we take every 12 frames in the training data set as a time segment to count the number of people's actions switchs between adjacent time segments, then we can get a actions  switching matrix, in which each unit represents the times of actions switching (includes self swich to self). Based on this raw statistical data, we can calculate the swiching probability by softmax function in each row.

\begin{figure}[h]
\begin{center}
\includegraphics[width=0.70\linewidth]{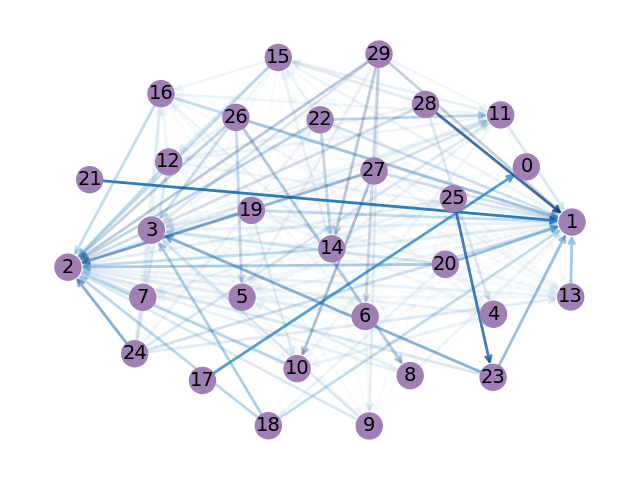}
\end{center}
   \caption{The weighted directed knowledge graph.}
\label{img7}
\end{figure}

Based on the probability switching matrix, we can build the knowlegde graph, which is shown in Figure.\ref{img7} (The Action-ID correspondence is list in Table.\ref{table0})

\begin{table}[h]
\begin{center}
\setlength{\tabcolsep}{0.8mm}{
\begin{tabular}{|c|c|c|c|c|c|}
\hline
\textbf{Action} & \textbf{ID} & \textbf{Action} & \textbf{ID} & \textbf{Action} & \textbf{ID}\\
\hline
BG & 0 & Talking & 10 & Talking phone &20\\
Walking & 1 & Transport & 11 & Tunning & 21\\
Standing & 2 & Unloading & 12 & PickUp & 22\\
Carrying & 3 & Pull & 13 & Using Tool & 23\\
Gesturing & 4 & Loading & 14 & SetDown & 24\\
Closing & 5 & Open Trunk & 15 & Crouching & 25\\
Opening & 6 & Closing Trunk & 16 & sitting & 26\\
Interacts & 7 & Riding & 17 & Object Transfer & 27\\
Exiting & 8 & Texting Phone & 18 & Push & 28\\
Entering & 9 & PP Interaction & 19 & PickUp & 29\\
\hline
\end{tabular}}
\end{center}
\caption{Correspondence between \textbf{Actions} and \textbf{IDs}.}
\label{table0}
\end{table}

\noindent\textbf{(2) Baseline method:}

We compare our method with the recent baseline: Next \cite{liang2019peeking} is an end-to-end model utilizing rich visual features about the human behavioral information and interaction with their surroundings to predict human actions, which is similar with our work in this paper. This method uses LSTMs to extract features, and directly splits different types of features into a long vector. And based on the conbined feature, they can get the prediction results.

\noindent\textbf{(3) Comparisons with baseline method: }

As this is a multi-lable classification problem, we use Mean Average Precision (mAP) \cite{sechidis2011stratification} to evaluate the results.

\begin{table}[h]
\begin{center}
\setlength{\tabcolsep}{0.8mm}{
\begin{tabular}{|l|c|c|c|}
\hline
Features & \textbf{Next}(mAP) & Our Model(mAP) & Improve \\
\hline\hline
Full Features & 0.192 & \textbf{0.213} & 10.9$\%$ \\
No P-Objects & 0.198 & \textbf{0.213} & 7.6$\%$ \\
No P-Scene & 0.206 & \textbf{0.211} & 2.4$\%$ \\
No P-Keypoint & 0.190 & \textbf{0.212} & 11.6$\%$ \\
No P-Appearence & 0.154 & \textbf{0.179} & 16.2$\%$ \\
\hline
\end{tabular}}
\end{center}
\caption{Comparison to baseline method \textbf{Next} on the ActEV/VIRAT dataset}
\end{table}

We test the two methods (Next and our method) with different input pre-extracted features, the experiment results are shown in Table.1. The column on the left shows the experimental results of Next, and the one on the right corresponds ours. From Table.1, we can find that our method has a 2.4$\%$ - 16.2$\%$ improvement under each condition.

After in-depth analysis into the Next’s results, we find that the P-Objects and P-Scene are noisy features, addition of which decrease the prediction accuracy of \textbf{Next}. Different from Next, our method is not sensitive to these noises. It is mainly because that our model uses attention modules and hierarchical aggregation mechanism to embed features, which makes it got the related features selection and strong noises anti-interference ability (the performace of our method with full features is 10.9$\%$ better than \textbf{Next}). 

\noindent\textbf{(4) Comparisons with different model types: }

In this part, we try to analysis the impacts of different parts of our model on the prediction results and the following three experiments are conducted. 

\begin{table}[h]
\begin{center}
\begin{tabular}{|l|c|c|}
\hline
Model Type & (mAP) \\
\hline\hline
Human nodes only & 0.194 \\
Human and Objects nodes & 0.208 \\
Human, Objects and Shadow nodes & \textbf{0.213} \\
\hline
\end{tabular}
\end{center}
\caption{Comparison results with different model types.}
\end{table}

First, we examine the impact of human interactions on the prediction and build the graph structure of spatial relationships by utilizing “human” information alone. The human-objects relationships as well as causal relationship between actions are not considered. In this case, the mAP of result is 0.194 (Table.2) which is similarly to that of \textbf{Next}. This means that a proper designed feature aggregation mechanism is important and is helpful to the action prediction by using just simple features.
 
 \begin{figure*}[h]
\begin{center}
\includegraphics[width=18cm]{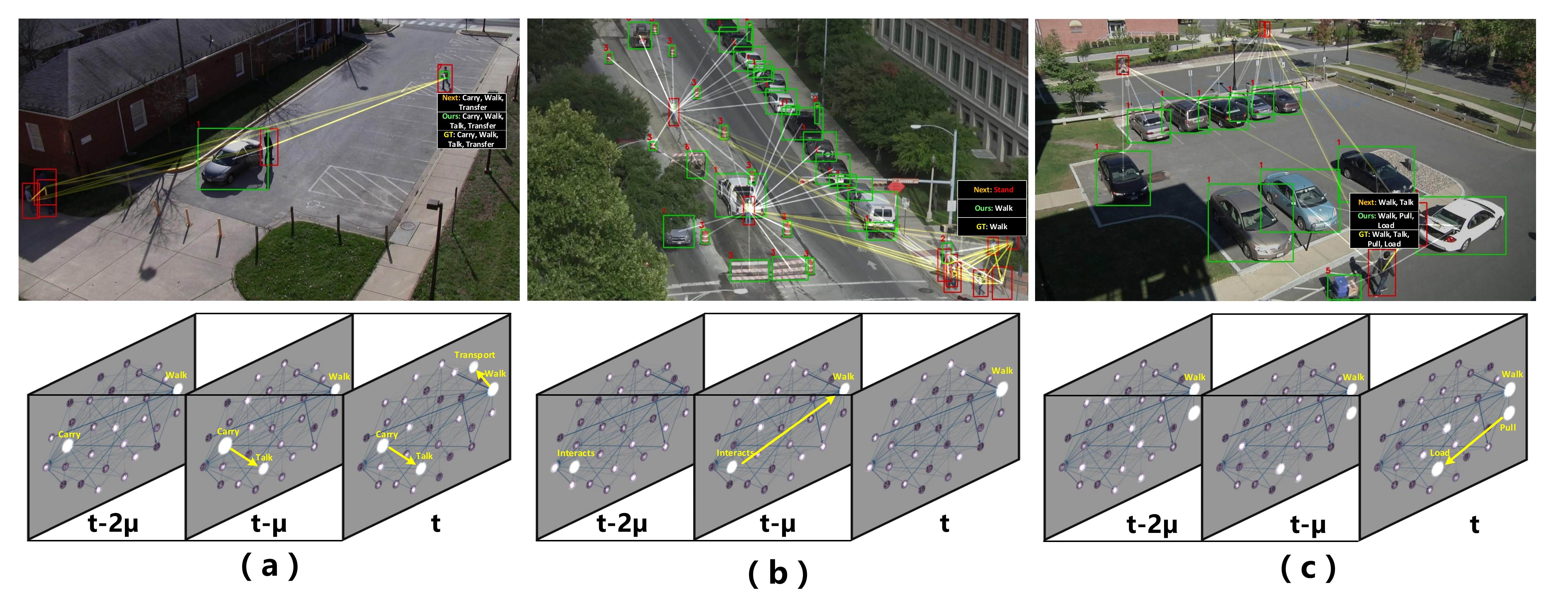}
\end{center}
   \caption{ The middle layer outputs visualization and comparison experiment results of our model.}
\label{img10}
\end{figure*}
 
Then we embed both the human and objects nodes into our model by the hierarchical attention aggregation mechanism. As shown from Table.2, the result is improved to 0.208 mAP, which implies that the introduction of human-objects relationships can provide useful information and enhance the ability of prediction model. Meanwhile, by comparing with the results of Next, we can find that directly introduce the scene information may produce noisy features, which will decrease the prediction performance. On the other hand, by modeling the action-scene relationship with our hierarchical graph attention module, the features of different scene elements can be effectively aggregated, which significantly improves the results of actions prediction. 

Finally, we integrated the causality features into the heterogeneous graph of the action-scene relationship in the form of “shadow node”. Moreover, because of the designing of self-attention module, we can determine the time when the knowledge graph information should be activated. In this case, the prediction result acheives 0.213 mAP, which is 10.9$\%$ higher than Next. It means that introducing causal relationships between actions can further improve the prediction result of our method. 

\noindent\textbf{(5) Qualitative analysis: }

We compare the performance of our model outputs and the baseline while visualize the mid-level outputs of our model as shown in Figure.\ref{img10}. In spatial dimension, the yellow lines are relationships between humans, and the white represent relationships between human and objects. Brightness of the connecting line corresponds to the weight of attention. Accordingly, in the time dimension, the brightness of different action nodes means the probability of each action prediction output. 

Figure.\ref{img10}(a) shows a simple scenario, in which a person is carrying a box to another one near the car. The spatial relationships between human-scenes are visualized, from which we can see that the person with box pays more attention to the car and the driver ahead (the lines between them are brighter than others). For the visualization of casual relationships, nodes in the knowledge graph are sequentially activated from ''walking'' and ''carrying'' to ''talking'', ''transfer'', ''walking'' and ''carrying'' during the period [$t-2\mu$, $t$] which are consistent with human cognitions.

Figure.\ref{img10}(b) depicts a more complex scene and we take the person at the bottom right of the view as an example to illustrate the visualization performance of our method. As shown form Figure.\ref{img10}(b), spatial relationships between the person and the surrounding ones are closer than others. Meanwhile, it can be inferred that the person is more likely to cross the road with others together, which is inconsistent with ground truth and demonstrate the effectiveness of our method in the complex environment.

Figure.\ref{img10}(c) is a scene contains cooperation. We can find that the person with baggage pays simillar attentions to both the blue and white car, which may confuse the prediction result. But the transfer of high interaction intention of the partner in front enables our model to infer the correct answer, which illustrate the hierarchical graph attention module is useful to eliminate ambiguities. The result is consistent with the results obtained according to the causal reasoning model. 

\section{Conclusions}

In this work, considering the spatial and causal relational reasoning mechanism for action prediction in human beings, we proposed a spatial and causal relationship based graph reasoning network (SCR-Graph), which can be used to predict human actions by modeling the action-scene relationship and actions causal relationship in spatial and temporal dimensions, respectively. By introducing the hierarchical graph attention module in the spatial dimension, our model was able to fuse features of different kinds of scene elements using different strategies. In the temporal dimension, we designed a knowledge graph based causal reasoning module by sequential nodes activation process. Moreover, we proposed a fusion method for spatial and causal relationship features, with self-attention shadow nodes. The effectiveness of our method is evidenced by its favorable performance, as compared to existing methods. 

In the future, we plan to explore the method to build an effective relational reasoning model in the original video, without annotation bounding boxes. This is a difficult problem because in a complex environment, it is difficult for the existing human and object detection and tracking algorithms to achieve reliable results. Therefore, our next goal is to design a more adaptable relational reasoning model.

\textbf{Acknowledgement: } Research supported by the Program of National Natural Science Foundation of China (No.U1609210, No.91748208) and Natural Science Foundation of China-Joint Funds of Liaoning Province (No.U1508208).

{\small
\bibliographystyle{unsrt}
\bibliography{egbib}
}

\end{document}